\documentclass[journal, twocolumn]{IEEEtran}
\IEEEoverridecommandlockouts
\usepackage{ulem}
\usepackage[english]{babel}
\usepackage[utf8]{inputenc}
\usepackage[ruled,lined]{algorithm2e}
\usepackage{wrapfig, blindtext}
\usepackage{float}
\usepackage[noend]{algpseudocode}
\usepackage[letterpaper, margin=0.75in]{geometry}
\usepackage[font=scriptsize]{caption}
\usepackage{url}
\usepackage[dvipsnames]{xcolor}
\usepackage{amsfonts}
\usepackage{amssymb,amsmath}
\usepackage{booktabs}
\usepackage{graphicx}
\usepackage{cases}
\usepackage{comment}
\usepackage{multirow}
\usepackage{tikz}
\usepackage{epstopdf}
\usepackage[colorlinks=true, allcolors=blue]{hyperref}

\usepackage[caption=false, skip=0pt, font=scriptsize]{subfig}
\usepackage{lipsum}
\usepackage{xargs}
\usepackage{gensymb}
\usepackage{mathtools,extarrows}
\newcommand{\norm}[1]{\left\lVert#1\right\rVert}
\usepackage[square,numbers]{natbib}
\title{\bf 
 \LARGE
\vspace{6 mm}Far-UVC Disinfection with Robotic Mobile Manipulator}

\author{
Ishaan Mehta$^{\dagger}$,
Hao-Ya Hsueh$^{\dagger}$,
Nikolaos Kourtzanidis$^{\dagger}$, 
Mateusz Brylka$^{\dagger}$  and 
Sajad Saeedi$^{\dagger}$

\thanks{$^{\dagger}$Ryerson University, Dept. of Mechanical and Industrial Engineering, \textcolor{white}{Xx}Email: {\{ishaan.mehta, hhsueh, nkourtza, mbrylka, s.saeedi\}@ryerson.ca}}%
}

\begin{document}
\maketitle



\begin{abstract}

\textcolor{black}{The COVID-19 pandemic has demonstrated the need for a more effective and efficient disinfection approach to combat infectious diseases. Ultraviolet germicidal irradiation (UVGI) is a proven mean for disinfection and sterilization and has been integrated into handheld devices and autonomous mobile robots. Existing UVGI robots which are commonly equipped with uncovered lamps that emit intense ultraviolet radiation suffer from: inability to be used in human presence, shadowing of objects, and long disinfection time.} \textcolor{black}{These robots also have a high operational cost.} \textcolor{black}{This paper introduces a cost effective 
germicidal system that utilizes UVGI to disinfect pathogens, such as viruses, bacteria, and fungi, on high contact surfaces (e.g. doors and tables). This system is composed of a team of 5-DOF mobile manipulators with end-effectors that are equipped with far-UVC excimer lamps. The design of the system is discussed with emphasis on path planning, coverage planning, and scene understanding. 
Evaluations of the UVGI system using simulations and irradiance models are also included. Please see the project's website for videos and simulations of the robot.
\footnote{\href{https://sites.google.com/view/g-robot}{https://sites.google.com/view/g-robot}}}

\end{abstract}

\hspace{-2 mm}

\section{Introduction}

\textcolor{black}{The onset of the novel coronavirus disease (COVID-19) has led to an increased focus on the development and need for more effective and efficient disinfection procedures that disinfect air and surfaces while minimizing unwanted contact between humans. Improvements for disinfection practices that can be incorporated into daily life could potentially lower both the transmission of various infectious diseases and the cost for disinfection.} 
\par
\textcolor{black}{A decontamination process is composed of both cleaning and disinfection. Cleaning is the physical removal of bio-burden, while disinfection is the elimination of infection causing pathogens (i.e. bacteria, viruses, prions, fungi, and other microorganisms)~\cite{de2007cleaning}. The conventional method for decontaminating often includes vacuuming, scrubbing, mopping and disinfecting with chemical disinfectants, such as sodium hypochlorite (bleach) and hydrogen peroxide. These manual approaches can be laborious, time-consuming, and error-prone, due to inconsistencies between human operators. Residue from chemical disinfectants can also be toxic and corrosive.}
\par
\textcolor{black}{On the other hand, ultraviolet radiation exposure can be used as a no-contact decontamination method. There is a wealth of evidence indicating the effectiveness of ultraviolet (UV) germicidal irradiation as a disinfection and sterilization approach for the prevention of various infectious diseases, including COVID-19, influenza, and tuberculosis~\cite{UVCREF1}. Short-wavelength ultraviolet light, known as UVC (200–280 nm), disrupts DNA/RNA of micro-organisms and terminates cellular activities and reproductions~\cite{UVbook}.} 

\par
\textcolor{black}{With this pandemic, there has been a rising interest in the use of UV carts/robots for disinfection to limit the spread of infections. These devices have been around from before the COVID-19 pandemic and were primarily developed to mitigate the spread of Hospital acquired infections (HAIs). Most devices are essentially uncovered UV lamps mounted on top of a manually operated or a robotic mobile base. There are two main challenges associated with these devices. \textcolor{black}{Firstly, UV exposure to humans could potentially lead to varying degrees of erythema, photoaging, skin cancer, and vision loss \cite{matsumura2004toxic,raeiszadeh2020critical}; therefore, these robots are only applicable when there is no human present.} Another issue is the high operating cost of these devices. They typically use an array of high-powered UV lamps which result in high energy consumption.}
\par
\textcolor{black}{To address these operational challenges, this work presents a novel design named Germicidal Robot (G-robot), in which a mobile manipulator is developed to disinfect target areas. Major contributions of this work are:} 

\begin{itemize}
    \item \textbf{1)~Automation:} No human intervention is needed during disinfection, while the system can interact with humans efficiently.

    \item \textbf{2)~Human Safety:} Using a robotic arm allows us to isolate the UV rays from direct contact with human skin. Our robot also uses far-UVC which has been proven to be safe for human skin. This enables our robots to operate in human presence.

    \item \textbf{3)~Improved Disinfection Range:} \textcolor{black}{By using a manipulator for UV disinfection, high-touch surfaces (e.g. door knobs, table tops, and light switches) can be disinfected more effectively than existing robots/devices. Areas that are cluttered and shadowed are also able to be reached more easily with the improved range of motion.}

    \item \textbf{4)~Cost Efficiency:} The design of our robot is modular and is based on low-cost open-source robotic platforms. In addition, the UV lights used have lower energy consumption leading to a system with less initial and operating cost.

    \item \textbf{5)~Resource Management:} Since our robot can be used alongside humans, it does not prevent the use of spaces being disinfected, such as patient rooms, labs, hallways etc. The safeness of far-UVC lamps also eliminates the need for additional oversight by staff.  

    \item \textbf{6)~Applications:} The proposed system can be deployed in various environments, including care facilities, clinics, animal farms, and schools.
\end{itemize}

\par
\textcolor{black}{This paper is organized as follows: Section \ref{sec:related_wrk} presents a summary of commercially available UV devices/robots and the ongoing research. Section 
\ref{sec:mech} presents the hardware overview of our design, and Section \ref{sec:soft} presents the autonomy stack of the robot. The experiments are presented in Section \ref{sec:expm}. Finally, future works and conclusions are included in Section \ref{sec:conc}}

\section{Related Work} \label{sec:related_wrk}
\vspace{-2 mm}
\textcolor{black}{UV disinfection systems have been used for various applications \cite{uvgireview4, raeiszadeh2020critical}. There is a rising interest in the use of area disinfection portable UV devices and robots. Portable UV devices typically consist of high power pulsed xenon lamps or an array of low pressure mercury lamps, and motion detectors \cite{UVCREF3}. The irradiance envelope of these lamps can disinfect a volume of space in all directions. These are used as a supplement to disinfection procedures in health care settings. Due to adverse effects of UV to human skin, these devices are used in empty rooms. The exposure to humans is further limited by covering of exposed windows. The motion-detection sensors also pause UV light emission when humans are detected. \textcolor{black}{Most of these devices are placed and maneuvered manually around set locations within a room by human operators \cite{UVCREF3}.}} 
\par
\textcolor{black}{UV robots improve on these devices in which the UV lights are mounted on a mobile base that offers autonomous capabilities \cite{UVCREF8}. These robots use simultaneous localization and mapping (SLAM) algorithms \cite{8436423} to build the map of the environment and navigate the environment autonomously to deliver UV dosages~\cite{ackerman2020autonomous}. Various companies are commercially offering portable UV devices and UV robots. Further studies have been done by these companies to demonstrate the efficacy of their products. The Xenex Lightstrike robot significantly reduced SARS-CoV-2 on hard surfaces and N95 respirators \cite{UVCREF9}. Tru-D Smart UVC robots reduced counts of pathogens, such as C. difficile and Acinetobacter in patient rooms at a tertiary acute care hospital  \cite{trud1}. Blue Ocean robotics's autonomous disinfection robot claims to disinfect rooms within 10 minutes. Their robot is being used in various hospitals in Romania, Croatia, and Italy \cite{uvdrobot}. Surfacide's Helios UV-C Disinfection System uses multiple robots to overcome shadowing issues. Their robots are being used by many hospitals across USA \cite{helio}. On the other hand, Honeywell's UV Treatment System is used to disinfect aircraft cabins and are being used by Qatar Airways. Their system can disinfect 30 rows of seating in about 10 minutes~\cite{honeywell}}.
\par
Current generation of UV robots have some disadvantages. Robots with uncovered UV lamps cannot be used in human presence. A room being disinfected by UV robots cannot be used. This is a major limiting factor for institutions, like hospitals, where there is a high demand of rooms due to constant influx of patients. Staff also has to ensure that there is no one present in the room and all the windows of that room are covered to prevent potential UV exposure. In addition, many high touch surfaces might not get disinfected properly due to occlusion by other objects and shadows.
\par
\textcolor{black}{Due to increasing demand in cleaning and disinfecting, several works are underway to seek improvements in germicidal and sanitation technologies, including UV robots. The use of manipulators for UV disinfection was proposed in  \cite{manipulator}. This would enable sufficient disinfection of hard to reach surfaces. McGinn \textit{et al.} \cite{mcginn2020exploring} used a novel design to shield the radiation envelope of UV robots so that it could be used along with humans. Similarly, a prototype of autonomous UV disinfection robot with improvements in mechanical and electrical design was developed in \cite{ultrabot}.} 


\begin{figure}[t!]
\centering
\includegraphics[width=0.22 \textwidth]{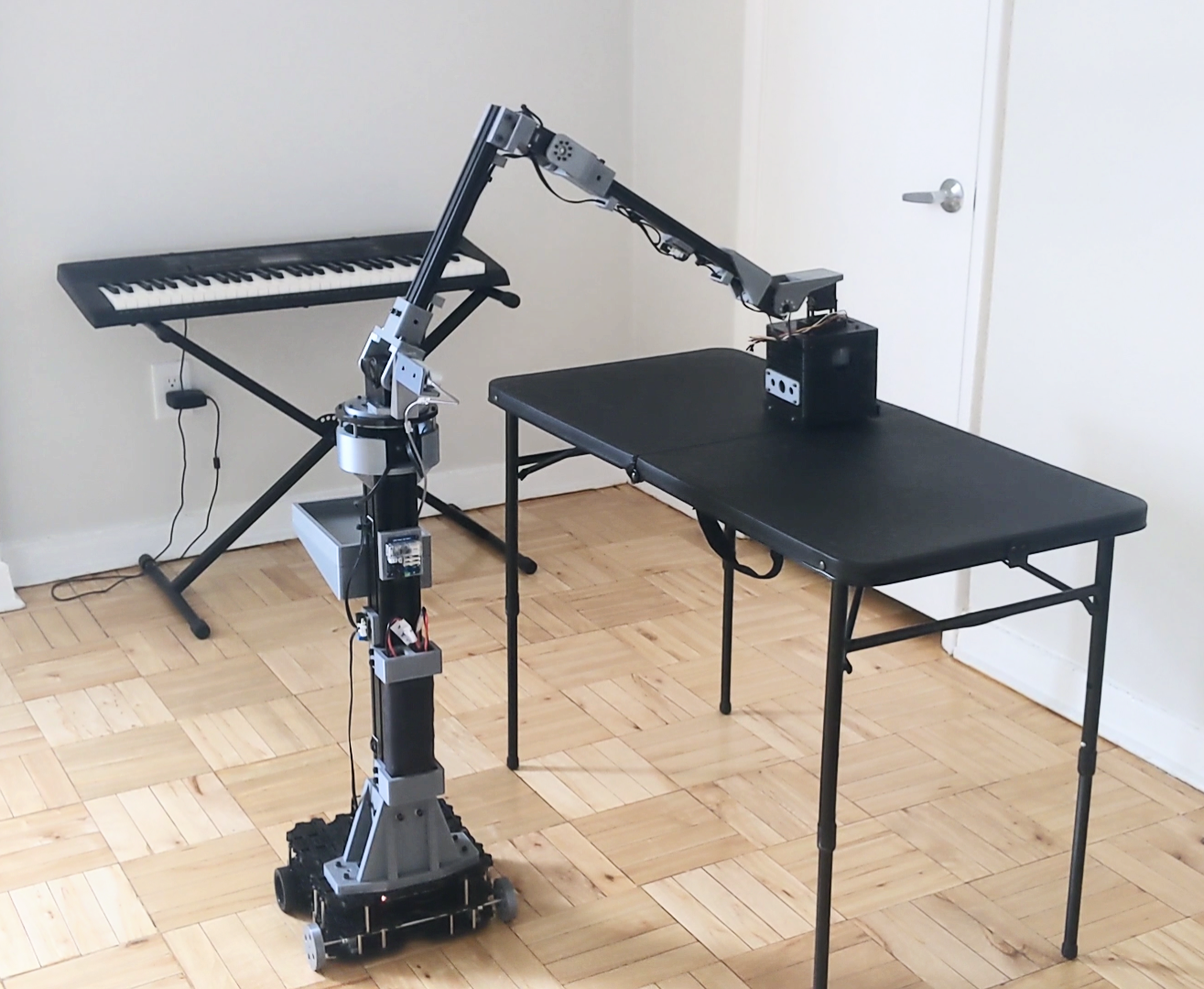}\\
\scriptsize{(a) G-robot santizing a table}

\centering
\includegraphics[width=0.35 \textwidth]{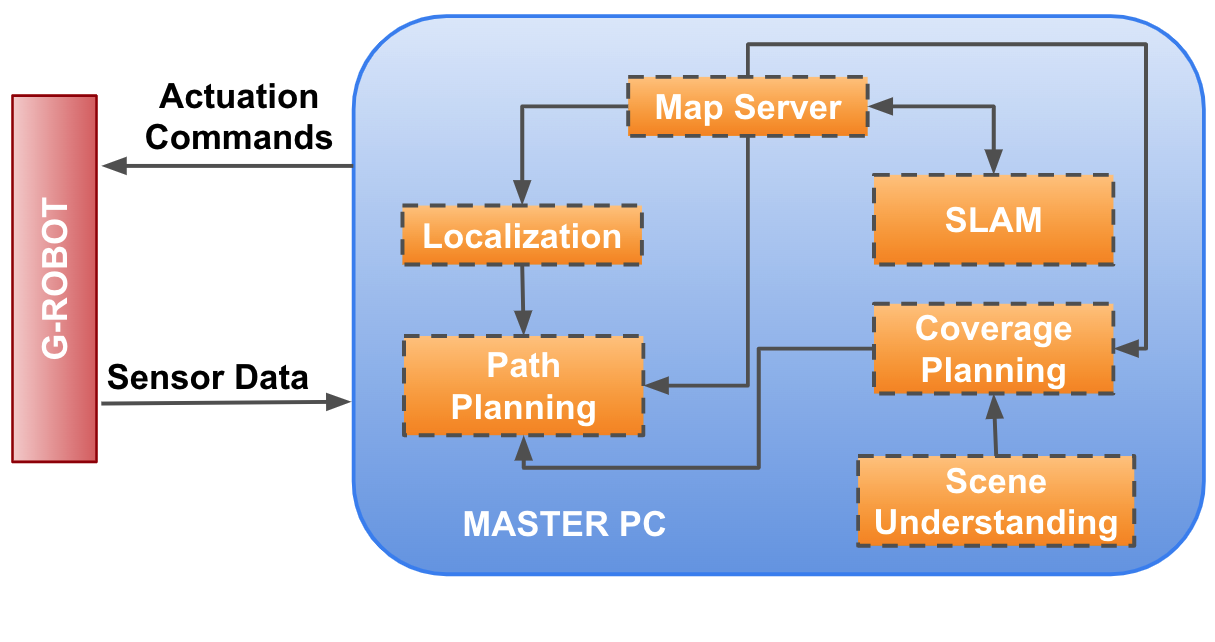}\\
\scriptsize{(b) Schematic representation of the software framework for G-robot} 
\caption{(a) The G-robot hardware is consisted of a mobile base with a 5-DOF robotic arm, where the far-UVC emitter is mounted on the end-effector. (b) Various modules in the software stack process the input data and generate actuator commands for the robot \label{fig:astack}}
\end{figure}

\vspace{-5 mm}
\section{Design Overview}\label{sec:mech}
\par
In this section, we review the motivation behind G-robot and its mechanical design.

\subsection{Motivation of Design}
\par
\textcolor{black}{UVC light sources come in various form-factors ranging from tube lights, excimer lamps to LEDs. Disinfection ability of UVC is dependent on its intensity. The intensity of the light decreases due to attenuation and dissipation (i.e. regions further away from the source of the light have lower intensity). Existing UV robots typically use an array of low pressure mercury lamps or pulsed xenon light and they require high power consumption.}
\par
\textcolor{black}{One of the main motivations for adopting a mobile manipulator design is based on the fact that a sufficiently strong disinfection dose can be achieved by placing the UVC light source very close to the target surface. With such a configuration, a low powered UV light could be used to get high disinfection rates in a shorter period of time. In order to demonstrate this point, we present a simulated experiment in Section~\ref{sec:expm}-A. We show that the manipulator design is able to generate strong doses with a relatively lower power consumption. Mathematically, the UV dose, $D$ [$J/m^2$], for pathogens subjected to UV radiation is given by \cite{UVbook}: }

\small
{\begin{equation} \label{eq:uvdose}
D = E_t \cdot I_r, \end{equation}}\normalsize
\normalsize
\noindent where $E_t$ is the exposure time in seconds and $I_r$ is the irradiance of the light source with SI units [$W/m^2$]. Different pathogens have varied susceptibility to UV radiation; hence, the required UV dose for an application should be selected based on the targeted microbes.

\vspace{-5 mm}
\subsection{Mechanical Design}

\par
\textcolor{black}{The mechanical design of G-robot, shown in Fig.~\ref{fig:astack}-(a), consists of two major physical systems: a wheeled mobile base and a 5-DOF robotic arm. The mobile base is used to transport the robotic arm to areas that require disinfection. The end effector of the arm is equipped with a far-UVC light source. The robotic arm then performs a sequence of movements to orient the end effector in optimal positions for disinfection of desired areas. In order to leverage existing open source projects, we implement a modified version of Turtlebot3 Waffle Pi \cite{tb3} developed by Robotis for the mobile base. The robotic arm manipulator is a custom design partially inspired by Trossen Robotic's ViperX 300 \cite{vpr}. 
}
\par
\textcolor{black}{ The UV module on the manipulator directly faces the target surface which reduces the risk of UVC exposure to humans. To further enhance human safety, our design includes the Ushio 12W Care222 far-UVC emitters for disinfection. Far-UVC light ($207-222 \, nm$) effectively inactivates pathogens, while being safe for human exposure. This is primarily due to the limited penetration distance of far-UVC from the outer layer of the mammalian skin \cite{FarUVCskin}.}

\section{Autonomy Stack}\label{sec:soft}
\par
\textcolor{black}{The robot has a sensor suite consisting of a stereo camera, LIDAR, and IMU to enable autonomy. Our framework utilizes a master-slave architecture, the robot is a slave that connects to a master PC via local area network. The communication is carried out through the subscriber and publisher framework in ROS \cite{o2014gentle}. The master PC is responsible for running all modules that enable autonomy. On the other hand, the slave is primarily responsible for communicating with the microcontroller and publishing the sensory data (e.g. images and laser scan) from the robot. } 
\par
\textcolor{black}{We have modified and adapted the standard ROS navigation stack \cite{nstack} for our software framework. A schematic representation and interaction of various modules in the software framework is shown in Fig~\ref{fig:astack}-(b). The mapping of the environment is performed through the SLAM module. This is achieved by traditional Grid-based SLAM (a.k.a gmapping) \cite{gmapros} with particle filtering, where raw laser data and odometry information is used to generate a grid map \cite{gmap1,gmap2}. This map is used by the localization module to determine the position of the robot in the map. The scene understanding module is responsible for detecting the target surfaces. These surfaces along with the grid based occupancy grid map of the environment are used by the coverage planner to generate waypoints for the robot. Finally, the path planning module determines the joint positions to execute the plans generated by the coverage planner. The details of some modules is presented next.}

\begin{figure}[t]
    \centering
    \includegraphics[width=0.45\textwidth]{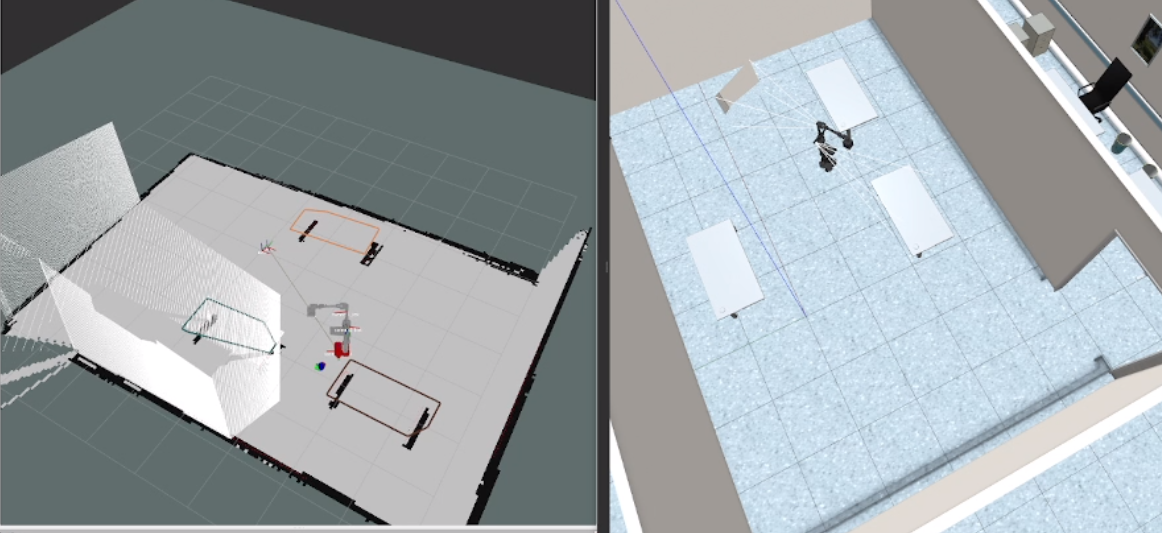}
    \caption{Plane Segmentation and Gmapping Of Proposed Model in Simulated Hospital Environment and in Rviz .}
    \label{fig:plane2}
\end{figure}

\subsection{SLAM and Scene Understanding}

\par

\par
\textcolor{black}{In order for the robot to have the ability to plan and execute its disinfection tasks autonomously, it is essential to build an understanding of the environment first. The SLAM module is used to build a map of the environment and the scene understanding module identifies surfaces that require disinfection within the mapped space.}

At the initial stage of our implementation, the main objective was to target tabletop surfaces. We adopt plane segmentation tools from \cite{Fallon2019PlaneSeg} to accomplish this goal. The original algorithm does not support utilities, such as map saving, updating, and association of the planes. We have augmented the original algorithm to include these features by adding a polygon map correction and updating feature. These utilities further enable us to save a dictionary of detected planar surfaces of interest that the robot can interact with. 
\par
The camera used for the received point cloud is the Intel RealSense Depth Camera D435i. The algorithm follows an 8-step procedure, which is depicted in Algorithm 1. In line 1, the preprocessing filter points only within $2.5 \, m$ away from mobile base and transforms the points with respect to the global map frame. In line 2 the processed point cloud is then further downsampled and the reduced number of points are input to the ground remover. Once the ground plane points are removed, the robust normal estimator in line 4 returns a list of estimated normal coefficients. In lines 5 and 6, incremental point by point checking occurs and the extracted labelled points on the hulls are then fit based on a maximum area scoring criteria. Lastly, polygon association and the updating of the map occurs in the final steps.

\begin{figure}[h]
    \centering
    \includegraphics[width=0.48\textwidth]{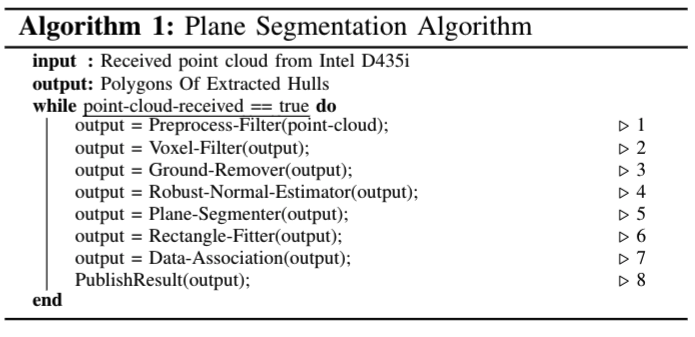}
\end{figure}



\normalsize
 Following the plane segmentation module, the final component of the procedure involves a rectangle fitter. The rectangle fitter handles the estimated planes and their corresponding labelled points, to produce a hull of each plane points based on a set objective. The objective chosen in this application was to return the hull of each plane whose points produce the largest area. The array of block fitted planes and their corresponding extracted hull points are stored in a Polygon Dictionary. As the robot navigates the desired area, the detected polygons are checked for relation with any current polygons in the dictionary, and if so, will be updated or swapped for the newly detected polygon that has met the association criteria and also has a larger area. The polygons are considered to be associated if their corresponding centers are within 1.3 metres away and at least one of the corresponding four corners are within 30 centimetres away. 

\par

\par
The plane segmentation module and Gmapping module are run alongside one another as shown in Fig.~\ref{fig:plane2}. The overall mapping consists of a 2D occupancy grid map and the estimated polygon hull points for each plane of interest.

\subsection{Path Planning}
\par
\textcolor{black}{ Once an understand of the disinfection environment and surfaces have been established, path planning on the mapped space can be conducted. The path planning module is responsible for motion planning and trajectory execution of the mobile base and robotic manipulator. The planning problem of our system is divided into two groups:}
\begin{itemize}

\item \textcolor{black}{\textbf{Mobile Base Planning:} As mentioned previously, we use turtlebot3 as our mobile base. Path planning is done by using the open source 2D navigation module where information from odometry, sensor streams, and goal pose, outputs velocity commands to the differential wheeled base \cite{navi}. This module is primarily responsible for navigating the robot towards all detected target surfaces based on the trajectories produced on previously built map.}

\item \textcolor{black}{\textbf{Manipulator Path Planning:} This module generates the joint paths for the 5DOF robotic manipulator along the target surface. The MoveIt framework is utilized for motion planning of the manipulator \cite{moveit}. \textcolor{black}{The framework combines planning scene generation, a path planning pipeline, and trajectoryexecution manager into a centralized movegroup ROS node.  With representation of the world and the current jointstate of the manipulator determined by MoveIt, motion planning requests with required constraints can be made. Sampling  path  planning  algorithms  from  the  integrated  motion  planning  library,  Open  Motion  Planning  Library(OMPL), are then used to formulate valid manipulator trajectories within MoveIt. The inverse kinematic solver of MoveIt is used to determine the joint angles to reach the desired location on the target surfaces during Cartesian space planning.} These joint angle trajectories are converted into commands for simulated/real actuators using a PID control loop feedback mechanism from the ROS Control package\cite{ros_control}. }

\begin{figure}[h]

\centering
\includegraphics[scale=0.25]{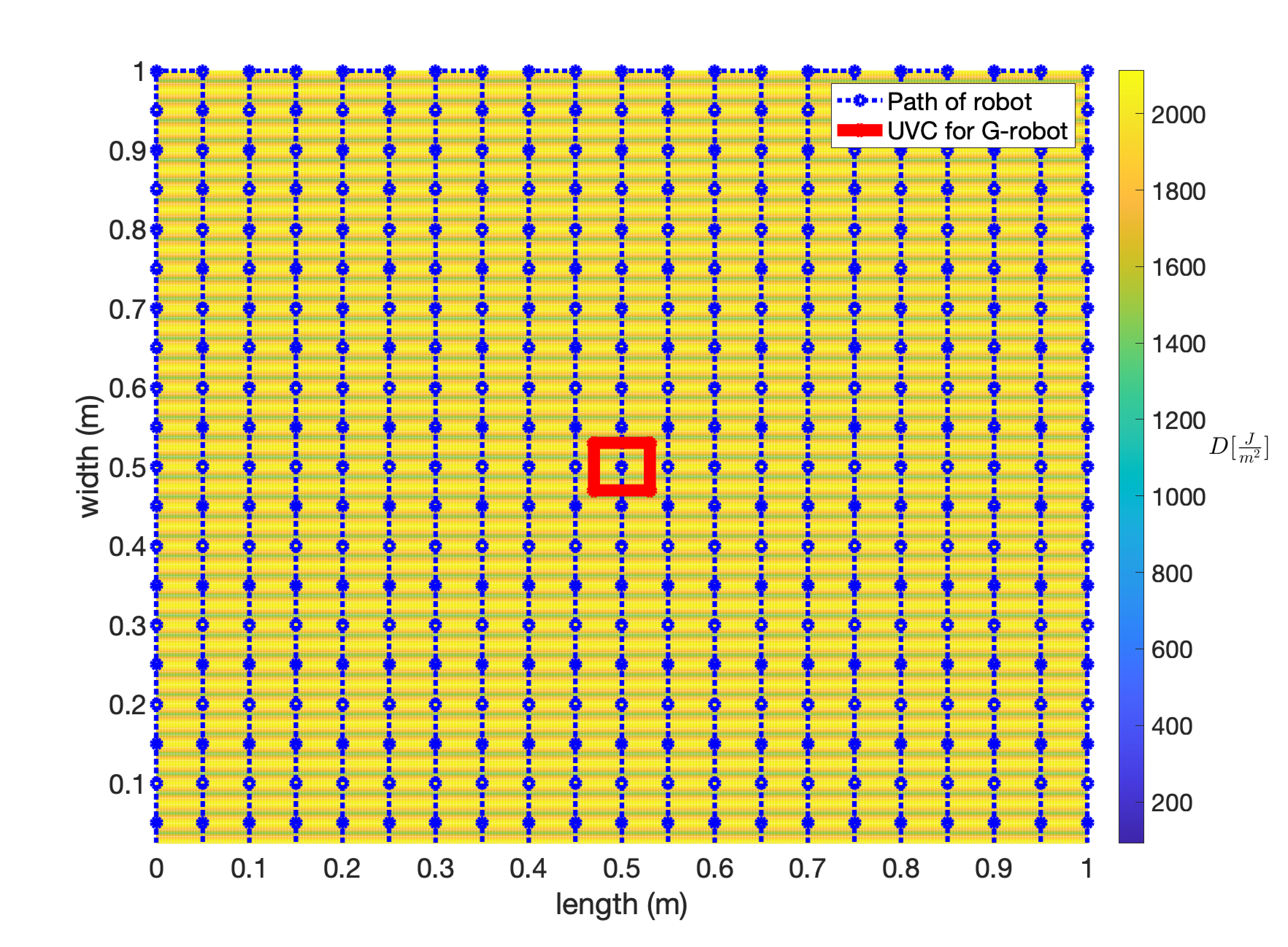} \\
\scriptsize{(a) Dosage map for G-Robot for wall and table-top}

\centering
\includegraphics[scale=0.25]{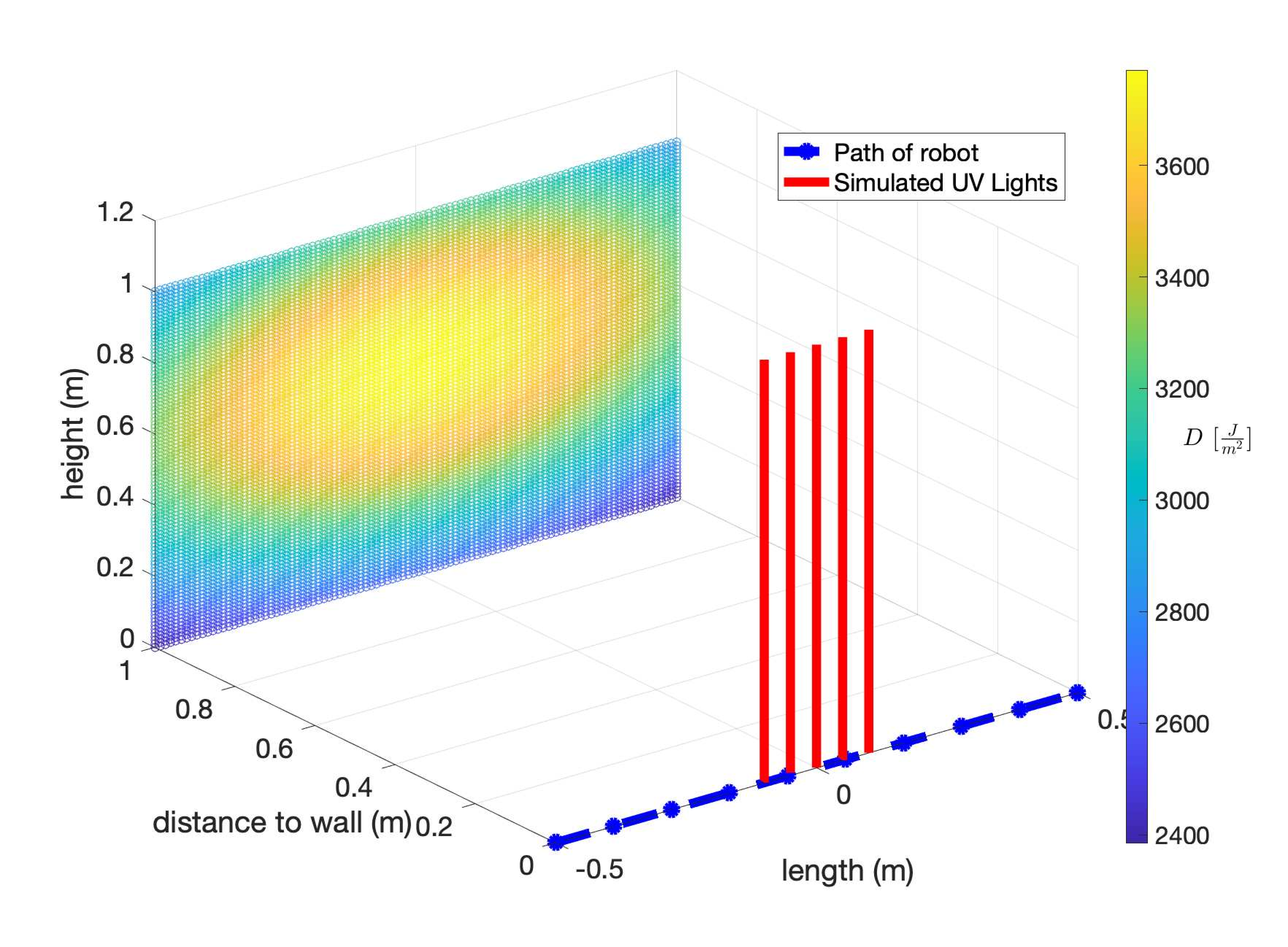} \\
\scriptsize{(b) Dosage Map for UV Robot for wall } \\

\centering
\includegraphics[scale=0.25]{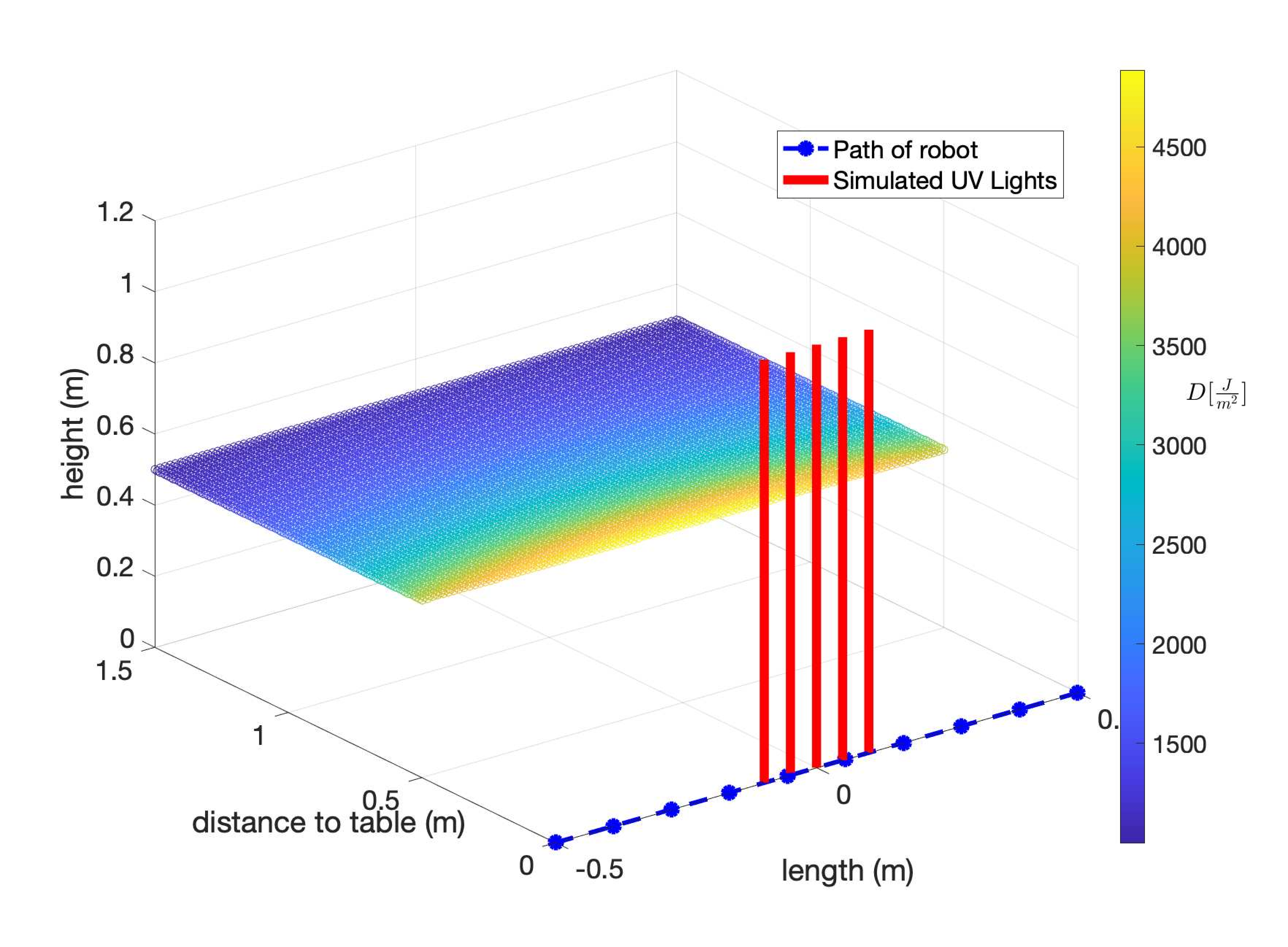} \\
\scriptsize{(b) Dosage Map for UV robot for table top } \\

\caption{ (a) G-robot uses a back and forth motion to scan a target surface. (b) \& (c) The UV robot simply moves along the straight line. Both robots are able to generate sufficient dosage required to inactivate harmful antibiotic resistant microorganisms, such as C.Diff and MRSE. The UV robot provides a much stronger dose because of high-powered lamps, but this comes at the cost of increased power consumption. Note that for all three cases, the light sources are highlighted in red at arbitrary locations. }\label{fig:dossageplot}
\end{figure}

\end{itemize}

\subsection{Coverage Planning} \label{sec:coverageplanning}
\par
\textcolor{black}{Coverage planning module is responsible for ensuring the complete disinfection of all the table-tops and other high touch surfaces in a given space. This is achieved in two steps
which are discussed next.}

\par
\textbf{Cell Allocation:} \textcolor{black}{A map generated by SLAM module is segmented into various regions called cells. For example, this segmentation is representative of various rooms, corridors and other spaces on a hospital or a school floor. The cell allocation module determines the visiting order of robot to these various cells. We solve this problem using Multi-Objective Travelling Salesman Problem (MOTSP). MOTSP finds a set of optimal tours for a single robot that are dependent on multiple objectives. These objectives can be conflicting in nature. In such cases, Pareto optimality is used to find the optimal solution set. MOTSP can be used for various real world scenario; for example, two criteria for our case could be the distance travelled by the robot and traversability of the path taken. }
\par
\textcolor{black}{Classical methods to solve MOTSP, such as non-dominated sorting genetic algorithm-II (NSGA-II) \cite{mtsp6} and multi-objective evolutionary algorithm (MOEA/D) \cite{mtsp7}, generate low quality solutions and are computationally expensive \cite{mtsp3}. There has been significant progress made in solving combinatorial optimization problems such as TSP through deep learning methods \cite{bengio2020machine,vesselinova2020learning,mazyavkina2020reinforcement}. Based on the success in single objective TSP, Li \textit{et al.} \cite{li2020deep} proposed deep reinforcement multi-objective optimization algorithm (DRL-MOA) to solve MOTSP. Their network used policy gradient. Firstly, the bi-objective optimization problem is converted into a single objective using the convex combination of the two objectives. Then, $N$ different networks are trained, each with different scalarizing factors, to approximate the Pareto front. We use DRL-MOA networks for our application. The motivation behind this choice can be demonstrated through an experiment for two objective MOTSP. The problem can be mathematically stated as:}

\small
\begin{equation}\label{eq:mob}
min_{\pi} \; \vec{F}(\pi) = (f_{1}(\pi),f_{2}(\pi))^\top\\
\end{equation}
\normalsize
\par 
\textcolor{black}{Here $\vec{F}(\pi)$ is a vector objective function and $\pi$  is a valid TSP tour. A TSP tour $\pi$ provides a sequence of visiting cities (i.e. cells/nodes) in the input graph. Each city is visited exactly once and finally the tour ends by returning  to the starting city. MOTSP is a multi-objective optimization problem that aims to find a set of Pareto optimal TSP tours $\Pi$ on a complete graph $G$. Here we consider the case of finding tours while optimizing for $2$ objectives. Each city in the input graph $G$ is a sequence of $n$ nodes in a four dimensional space $G = \{b_i^1, b_i^2\}_{i=1:n}$, where $b_i^m \in \mathbb{R}^2$ for each $m \in \{1,2\}$. The goal is to find a tour $\pi \in \Pi$ that visits each city once and can simultaneously optimize the objectives for $m \in \{1,2\}$: }

 \small
\begin{equation}\label{eq:tsp_obj}
     f_{m}(\pi|G) = \norm{b_{\pi(n)}^m -b_{\pi(1)}^m }_2 + \sum_{i=1}^{n-1} \norm{b_{\pi(i)}^m - b_{\pi(i+1)}^m }_2 
\end{equation}
\normalsize
\par
\begin{figure}[ht!]
\centering

\subfloat[200 City MOTSP]{\includegraphics[width=0.28\textwidth]{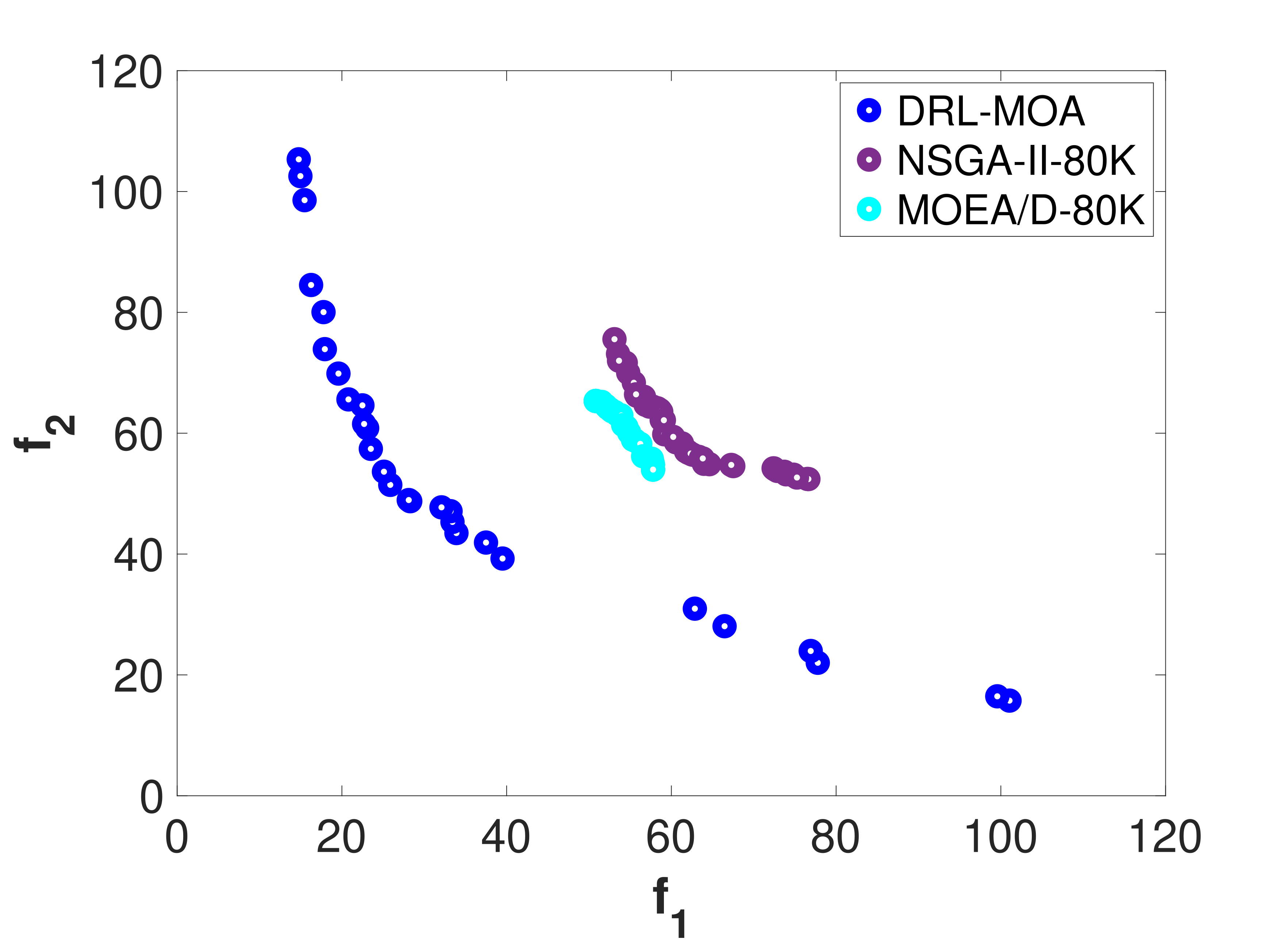}} 

\subfloat[1000 City MOTSP]{\includegraphics[width=0.28\textwidth]{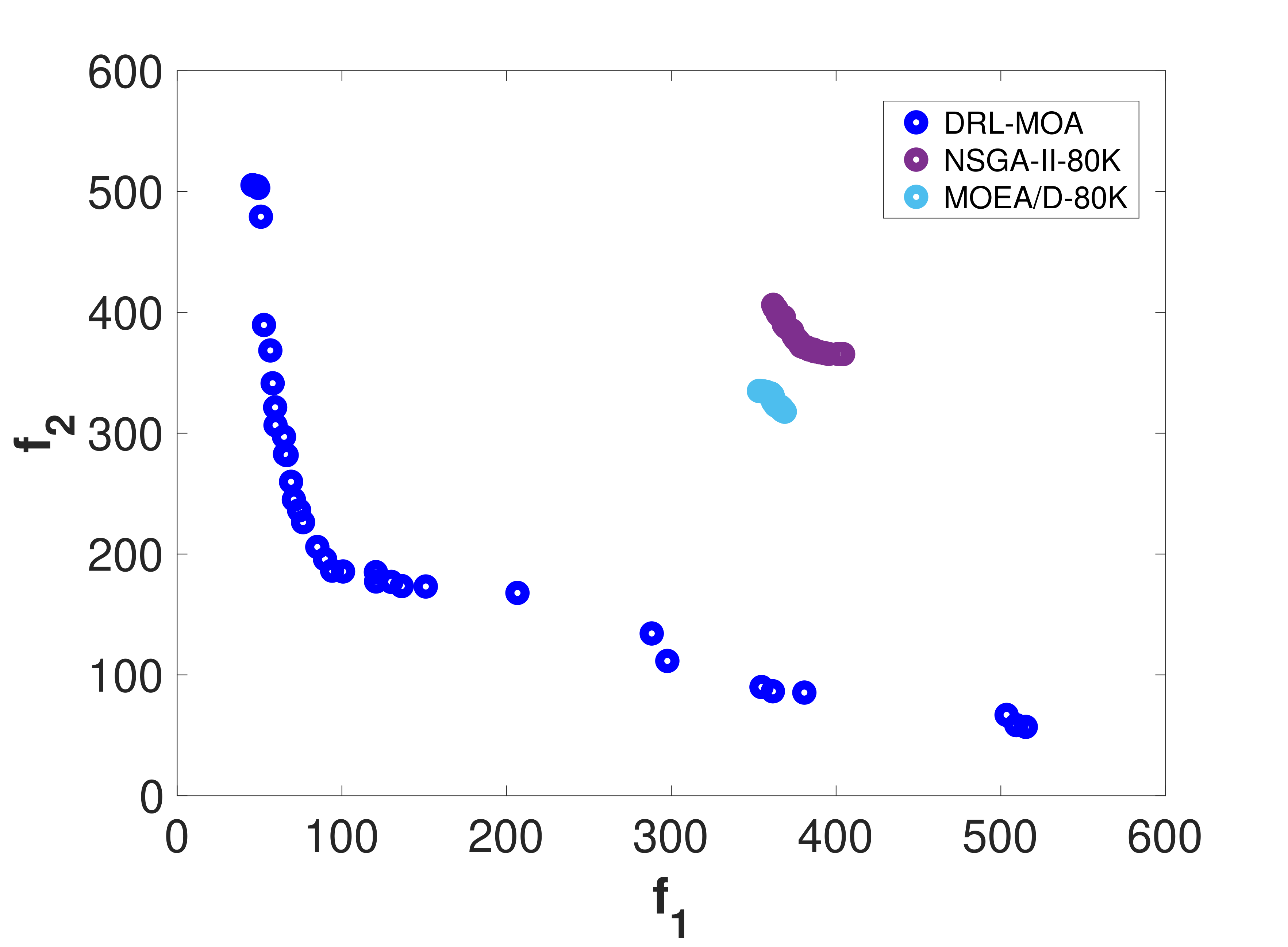}} \\

\caption{(a)-(b)Visualization of the dominant solutions for 2 objective MOTSP. It can be seen that DRL-MOA achieves significantly better objective values. \label{fig:Plots_panet}}
\end{figure}

\textcolor{black}{We tested the 2-objective MOTSP for 200 and 1000 cities respectively. DRL-MOA significantly outperforms classical approaches. Comparison of Pareto front generated by DRL-MOA and other methods for 2-objective MOTSP are shown in Fig.~\ref{fig:Plots_panet}-(a)\&(b). It can be clearly seen that DRL-MOA has significantly lower objective values than other algorithms.}
\par
\textcolor{black}{\textbf{Waypoint Generator:} Once a robot reaches its assigned location, the scene understanding module will identify various target surfaces such as desks, table tops, and doors. Based on the identified surfaces, a patterns will be generated for the manipulator planner to disinfect the complete surface. A simple disinfection pattern for a table-top is shown in Fig.~\ref{fig:dossageplot}-(a), where the path and waypoints are highlighted in blue.}

\vspace{-5 mm}
\section{Experiments} \label{sec:expm}
\par
In this section, we present two experiments to evaluate the efficacy of our system. The first experiment compares the disinfection dose generated by our robot and a standard UV robot in a simulated environment. The second experiment demonstrates our robot in action in real world and simulations.
\vspace{-1.5 mm}
\subsection{Dosage Comparison}
\par
\textcolor{black}{In this experiment, we compare the dosage map generated by a conventional UV robot and our robot, G-robot. The key details for the experiment are as follows:}
\begin{itemize}
    \item \textcolor{black}{\textbf{Light source:} The disinfection dosage for each robot is generated by simulating the irradiance profile of its light source:} \begin{itemize}
        \item \textcolor{black}{ To simulate UV robot, we model the irradiance profile of a set of five low pressure mercury lamps. Each lamp has a length of $1.19 \, m$ and a radiant power of $27 \, W$. In addition, each lamp has a wattage of $115 \, W$. We use view factor model \cite{UVbook} to generate dosgae profile of these lamps.}
        \item \textcolor{black}{For G-robot, we use a UVC LED panel made of 25 LEDs. The size of the LED panel is $25 \times 25 \, mm$. Each LED has a radiant power of $28 \, mW$. The wattage of the panel is $30 \, W$.}
    \end{itemize}
    \item \textcolor{black}{\textbf{Environment:} We compare the dosage and energy consumption of the robots for two cases. The first case represents a $1 \times 1 \, m$ wall and the second one represents a table of the same size at a height of $0.5 \, m$. For this test, each robot irradiates the target surface for 6 minutes. The dosage profile of G-robot in both cases is the same, because it primarily disinfects by getting close to the target surface (i.e. at a distance of $2 \, cm$ from the target surface). For typical UV Robots, we assume that the robot is at a distance of $1 \, m$ from the target surface.}
    \item \textcolor{black}{\textbf{Dosage Map:} Each robot moves along a certain trajectory, highlighted in dosage plots in Fig~\ref{fig:dossageplot} during the disinfection operation. The dosage map is computed by aggregating the dosage received by each point on the target surface from various positions and configurations of the robot. The dosage is calculated based on eq-\eqref{eq:uvdose}. }
\end{itemize}

\par 
\textcolor{black}{The dosage maps for G-robot and UV Robot are given in Fig.~\ref{fig:dossageplot}. Key data is summarized in Table 1. Both UV robot and G-robot generate the sufficient dosage required to inactivate harmful antibiotic resistant microorganisms, such as MRSA and C.Diff. It is clear that UV robot generates much higher average dosage as compared to G-robot; however, this comes at an increased cost of operation. From Table 1, it can be seen that G-robot provides a much higher Dosage Per Unit energy consumption (DPU). This is because G-robot disinfects from a very close distance, and higher dosage is achieved at lower energy consumption as a result. On the contrary, UV robots use high-powered lamps which provide strong doses at an increased energy consumption. The major drawback of UV-robots is that with such strong intensities of UVC, they cannot be utilized in applications involving human presence. G-robot avoids these issues through a low powered light source and a design that enables and allows for operating alongside human presence. In addition, G-robot's manipulator allows for disinfection of shadowed areas and tight spaces that cannot be reached by a conventional UV disinfection robot.} 

 \begin{table}[h]
 \begin{center}
    \caption{Summary of Dosage and Energy Consumption}
        \begin{tabular}{|c|c|c|c|} 
        \hline
        \scriptsize{Device} & \scriptsize{Avg. Dose [$J/m^2$]} & \scriptsize{Energy Consumption [$W \cdot hr$]} & \scriptsize{DPU} \\
        \hline
       \scriptsize{G-robot} & \scriptsize{1790} & \scriptsize{\bf 3} & \scriptsize{\bf 596.6}  \\  
       \scriptsize{UV Robot} & \scriptsize{\bf 2240} & \scriptsize{57.5} & \scriptsize{38.9}\\ 
     
        \hline
        \end{tabular}
        \label{tab:dose1}
    \end{center}
\end{table}

\vspace{-4 mm}
\subsection{Robotics Experiment}

\begin{figure}[h]
\centering

\subfloat[Fleet of G-robots Sanitizing target surfaces in a simulated environment]{\includegraphics[width=0.32\textwidth]{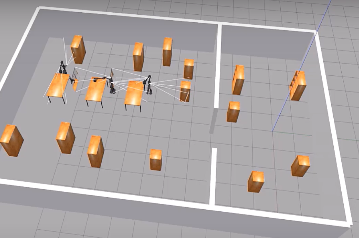}} 
\hfill
\subfloat[Fleet of G-robots sanitizing target surfaces]{\includegraphics[width=0.32\textwidth]{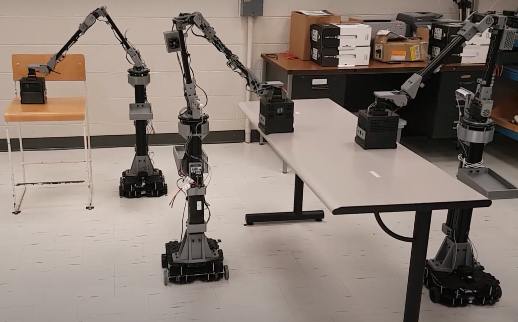}} \\

\caption{G-robots can be seen in action in simulated and real environments. In the simulated environment, the robots are seen disinfecting different target surfaces. In the real world environment, we demonstrate two robots disinfecting a table collaboratively. \label{fig:expm}}
\end{figure}

\par
We have tested G-robot in real and simulated environments. A visual representation of how the robots interpret the surroundings is shown in the simulated Gazebo environment of Fig.~\ref{fig:expm}-(a). In Fig.~\ref{fig:expm}-(b), we demonstrate a fleet of three robots working collaboratively to disinfect high touch surfaces in a classroom environment. In Fig.~\ref{fig:expm}-(b) the team of robots navigate to the target locations using the map produced by the SLAM and Scene Understanding module. The planned trajectories for disinfection are then generated by the coverage planner using the hull of extracted surface points detected.Finally, the planning module generates the joint angle commands for the simulated robots based on the data from the coverage planner. The robots are able to detect, navigate, generate, and execute the disinfection plans. 
\par
We also tested a fleet of our robots in a real world environment as shown in Fig.~\ref{fig:expm}-(b). In this case, two robots collaboratively disinfect a desk, whereas the third robot disinfects the back rest and seat of a chair. The entire operation was completed in a span of \textbf{1.5 minutes}. This further validates that our system can be deployed in real world situations to disinfect high touch surfaces. As briefly discussed in Section III.B, the inherent nature of our design and incorporation of the Ushio 12W Care222 far-UVC emitter, prevents detrimental effects on human skin and eyes. Our system can therefore be used alongside humans and is capable of providing disinfection around the clock. This is a significant advantage over existing UV robots of which can only be used in isolated/human-free environments.  
\par
Please refer to the website of the project to see visualizations of our simulated and real-world experiments, \href{https://sites.google.com/view/g-robot/home}{https://sites.google.com/view/g-robot/home}.

\vspace{-4 mm}
\section{Conclusions and Future work} \label{sec:conc}
\vspace{-1.5 mm}
In this work, we presented G-robot, a human-safe mobile manipulator used for UV disinfection. We demonstrated the efficacy of G-robot in terms of dosage distribution, energy consumption and its application in real world through our experiments. Our robot has many advantages over commercially available UVC robots, \textcolor{black}{including its ability to be used in human presence, and its improved disinfection effectiveness for cluttered and shadowed spaces.} Hence, G-robot can be deployed in hospitals to limit the spread of infectious diseases. Our future work will be focused on evaluating the effectiveness of G-robot in clinical settings. We also plan to use pixel level classification or semantic bounding boxes to further enhance the scene understanding. This would enable us to identify and generate paths for more complex surfaces in the environment. \textcolor{black}{Finally, a multi-robot system that facilitate collaboration of multiple G-robots for disinfection is also being explored.}  

\bibliographystyle{IEEEtranN}
\bibliography{bib} 
\end{document}